\pdfoutput=1

\documentclass[11pt]{article}

\usepackage[preprint]{acl}

\usepackage{times}
\usepackage{latexsym}

\usepackage{physics}
\usepackage{listings}
\usepackage{booktabs}

\usepackage[T1]{fontenc}

\usepackage[utf8]{inputenc}

\renewcommand{\cite}{\citep}
\renewcommand{\citeauthor}{\citealp}
\renewcommand{\citeyear}{\citeyearpar}

\usepackage{microtype}

\usepackage{inconsolata}

\usepackage[capitalise]{cleveref}

\lstdefinestyle{base}{
  language=C,
  emptylines=1,
  breaklines=true,
  basicstyle=\ttfamily\color{black},
  moredelim=**[is][\color{red}]{@}{@},
}

\usepackage{graphicx}

\title{Drop Dropout on Single-Epoch Language Model Pretraining}

\author{{Houjun Liu$^*$, John Bauer \and Christopher D.~Manning} \\
  Stanford University\\
  \texttt{$^*$houjun@stanford.edu}}

\begin{document}
\maketitle
\begin{abstract}
  Originally, dropout was seen as a breakthrough regularization technique that reduced overfitting and improved performance in almost all applications of deep learning by reducing overfitting. Yet, single-epoch pretraining tasks common to modern LLMs yield minimal overfitting, leading to dropout not being used for large LLMs. Nevertheless, no thorough empirical investigation has been done on the role of dropout in LM pretraining. Through experiments in single-epoch pretraining of both masked (BERT) and autoregressive (Pythia 160M and 1.4B) LMs with varying levels of dropout, we find that downstream performance in language modeling, morpho-syntax (BLiMP), question answering (SQuAD), and natural-language inference (MNLI) improves when dropout is not applied during pretraining. We additionally find that the recently-introduced ``early dropout'' also degrades performance over applying no dropout at all. We further investigate the models' editability, and find that models trained without dropout are more successful in gradient-based model editing (MEND) and equivalent in representation-based model editing (ReFT). Therefore, we advocate to \textbf{drop dropout} during single-epoch pretraining.
\end{abstract}

\section{Introduction}

Dropout \citep{hinton2012improving,srivastava2014dropout} is the method of randomly removing a certain percentage of features during each training pass. For the decade after the introduction of AlexNet \citep{krizhevsky2012imagenet}, the use of dropout became standard as a simple, highly effective regularization mechanism for very deep neural networks. Dropout helps create more robust feature representations, in particular, reducing feature co-adaptations \citep{hinton2012improving}, enabling the network to learn to make independent predictions from features and leading to more robust networks. 



Dropout was originally introduced at a large rate of $p=0.5$ \citep{hinton2012improving}, but the dropout rate used in NLP is steadily reducing in the decade that follows. The original transformer architecture \citep{vaswani_attention_2017} applies dropout $p=0.3$ at each of its network layers. BERT \citep{devlin-etal-2019-bert}, GPT-2 \citep{radford2019language}, and T5 \citep{raffel2020exploring} all used dropout $p=0.1$. Recent open language models such as LLaMA \citep{touvron2023llama} do not report any dropout use.

Alternate uses of dropout have emerged beyond regularization. \citet{liu2023dropout} highlights a novel use of ``early dropout'' as a stabilisation approach to reduce early underfitting. The authors found that applying dropout early reduces downstream underfitting, but this effect is reduced when dropout is applied throughout training. 


In our study, we examine the use of dropout, including early dropout, within the context of pretraining by investigating the effects of both standard and early dropout in pretrained language models. We pretrain both masked and decoder language models (MLMs and LMs), in particular, BERT-base \citep{devlin-etal-2019-bert} and Pythia 160M and 1.4B \citep{10.5555/3618408.3618510}, with varying levels of dropout at $p=0.0$, $p=0.1$, and $p=0.3$. Additionally, we apply early dropout (as in \citealt{liu2023dropout}) during the first $35\%$ of training; we then measure the downstream capabilities of these models at varying checkpoints. We measure morpho-syntactic understanding of linguistic minimal pairs (average BLiMP score) for decoder LMs and evaluate question answering for masked LMs. For all architectures, we additionally measure LM loss. We find that the complete removal of dropout (including early dropout) during pretraining yielded the most capable models across all measures. 

Recent investigations of LMs also show that their performance varies based on the degree of consistency with which they process inputs \citep{elazar-etal-2021-measuring}, pointing to the tension between having multiple distributed representations which dropout induces \citep{hinton1984distributed} and more localist approaches. We investigate this by measuring the means by which MLMs store factual knowledge through performing interventions on latent embeddings. By editing an MLM's embeddings through MEND \citep{mitchellfast}, a gradient-based model editing technique, as well as ReFT \citep{wu2024reft}, a representation-based model editing technique, we find that models trained without dropout can be more easily edited.

Finally, we discuss and contextualize the implications of this result. We release code and pretrained models.\footnote{https://github.com/Jemoka/dropfree}


\section{Related Work}
\paragraph{Dropout} The dropout mechanism \citep{hinton2012improving} has been extensively studied as a means to reduce feature co-adaptation, create ensembles, regularize, and improve gradient alignment \citep{srivastava2014dropout,baldi2013understanding,wager2013dropout,galDropoutBayesianApproximation2016,liu2023dropout}.

\paragraph{Knowledge Localization} We take the view that factual ``knowledge'' can be stored and elicited from neural networks \citep{petroni-etal-2019-language}, specifically, in the Multi-Layer Perceptron (MLP) after each layer's attention block \citep{geva-etal-2021-transformer}. Methods exist to measure the correctness and consistency \citep{elazar-etal-2021-measuring} of the stored knowledge and to probe for their stored location in terms of MLP activations \citep{dai-etal-2022-knowledge} as well as MLP parameters \citep{csordas2020neural}.

\paragraph{Factual Editing} Methods exist to edit ``knowledge'' stored in LMs, including tuning a low-rank subspace of the network \citep{hulora}, learning a parameter update through a surrogate network \citep{de-cao-etal-2021-editing}, projecting tuning gradients into edits \citep{mitchellfast}, probing orthogonal subspaces of representations \citep{wu2024reft}, or causal methods specific to mutating information flow in decoder models \citep{meng2022locating}.

\section{Approach}

\begin{figure*}[ht]
    \centering
    \hspace{-0.01pt}
    \begin{tabular}{ccc}
      \includegraphics[width=0.31\textwidth,trim={0.7cm 0.7cm 0.7cm 0.7cm},clip]{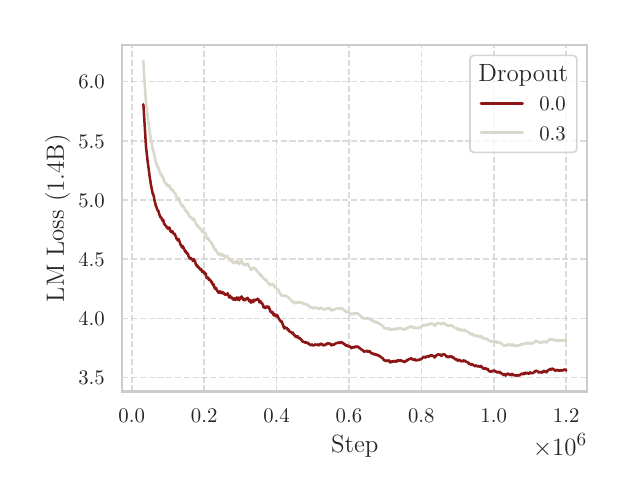} &
      \includegraphics[width=0.31\textwidth,trim={0.7cm 0.7cm 0.7cm 0.7cm},clip]{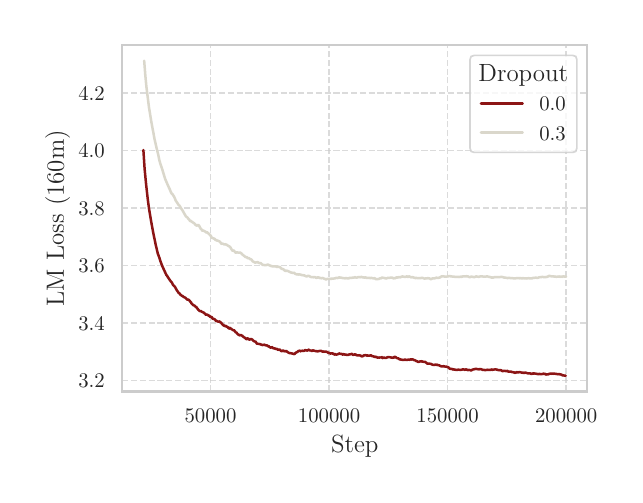} &
      \includegraphics[width=0.31\textwidth,trim={0.7cm 0.7cm 0.7cm 0.7cm},clip]{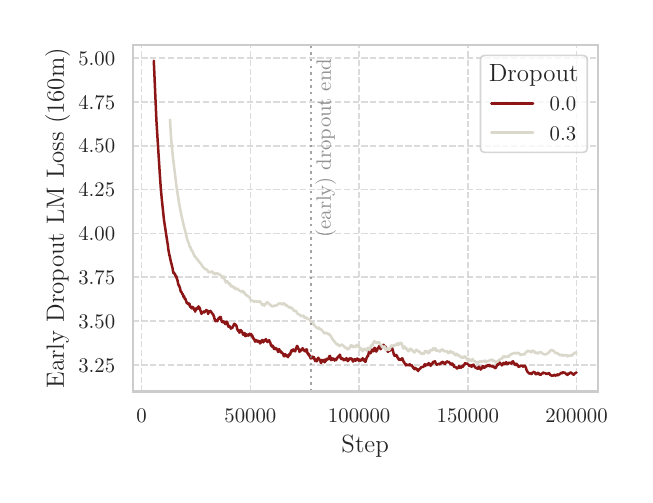} \\ 
      \includegraphics[width=0.31\textwidth,trim={0.7cm 0.7cm 0.7cm 0.7cm},clip]{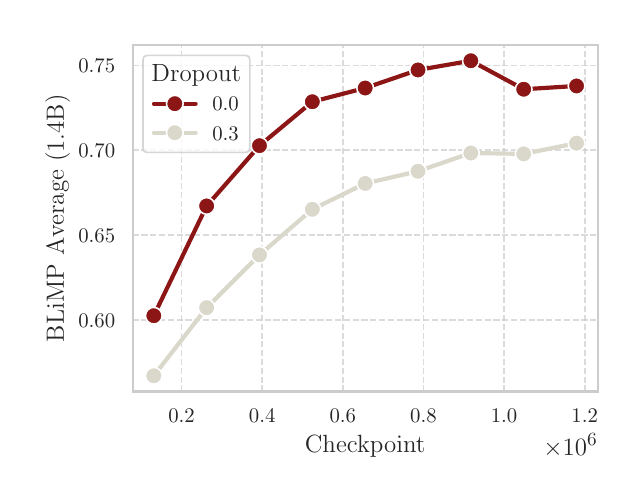} &
      \includegraphics[width=0.31\textwidth,trim={0.7cm 0.7cm 0.7cm 0.7cm},clip]{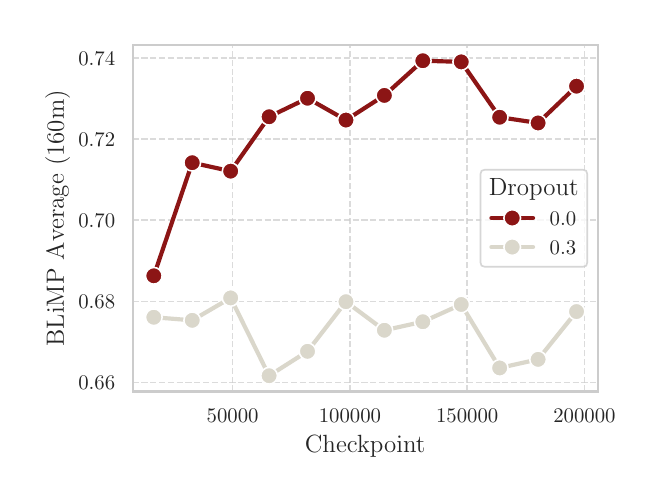} &
      \includegraphics[width=0.31\textwidth,trim={0.7cm 0.7cm 0.7cm 0.7cm},clip]{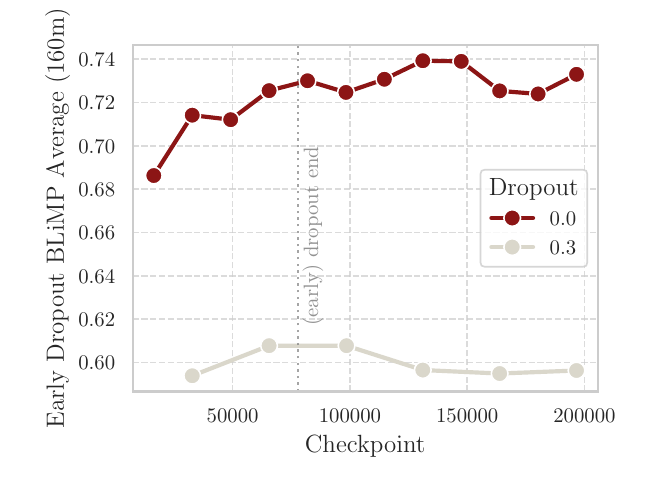}\\
        \includegraphics[width=0.31\textwidth,trim={0.7cm 0.7cm 0.7cm 0.7cm},clip]{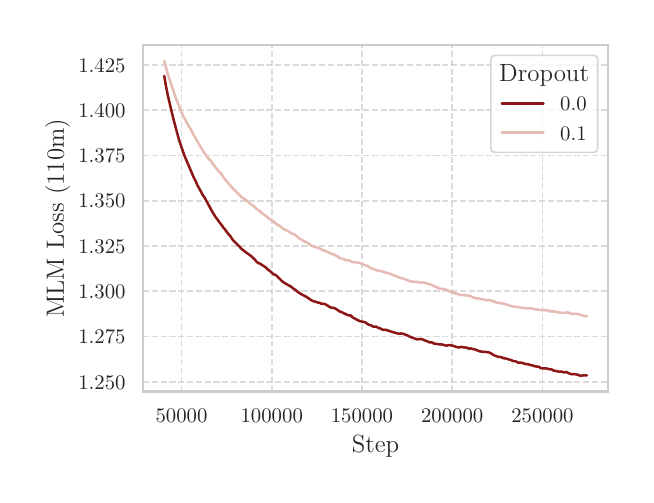} &
        \includegraphics[width=0.31\textwidth,trim={0.7cm 0.7cm 0.7cm 0.7cm},clip]{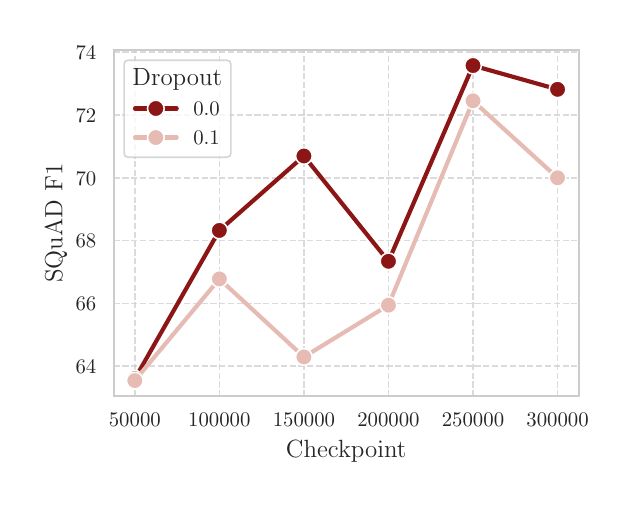}&
        \includegraphics[width=0.31\textwidth,trim={0.7cm 0.7cm 0.7cm 0.7cm},clip]{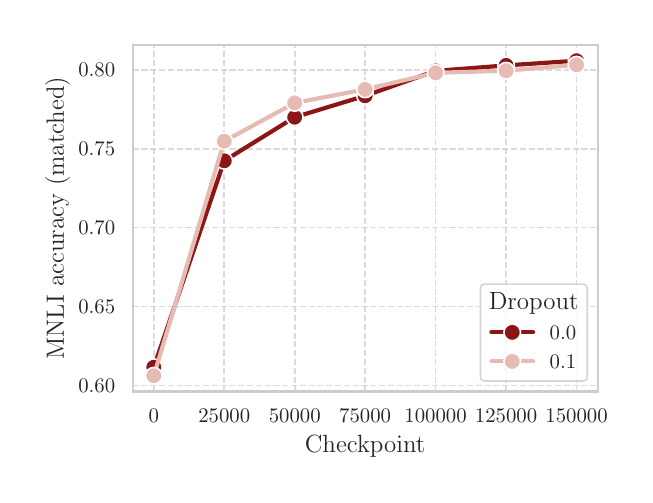}
    \end{tabular}

    \caption{top: language modeling loss with dropout $p=0.0$, $p=0.3$, and early dropout for decoder LMs; middle: mean BLiMP scores for these models; bottom: masked language modeling loss with dropout $p=0.0$, $p=0.1$, SQuAD F1 (answerable) dev-set scores, and matched MNLI scores.}
    \label{fig:large-figure}
\end{figure*}

\begin{figure}[ht]
    \centering
    \hspace{-0.01pt}
    \begin{tabular}{c}
      \includegraphics[width=0.45\textwidth,trim={0.7cm 0.7cm 0.7cm 0.7cm},clip]{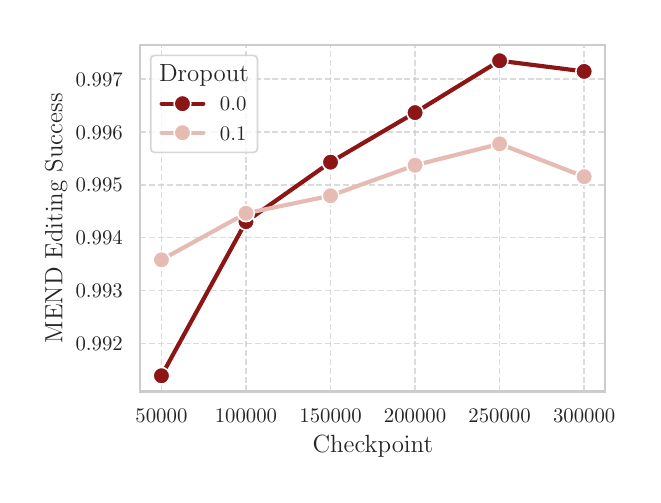}  \\
      \includegraphics[width=0.45\textwidth,trim={0.7cm 0.7cm 0.7cm 0.7cm},clip]{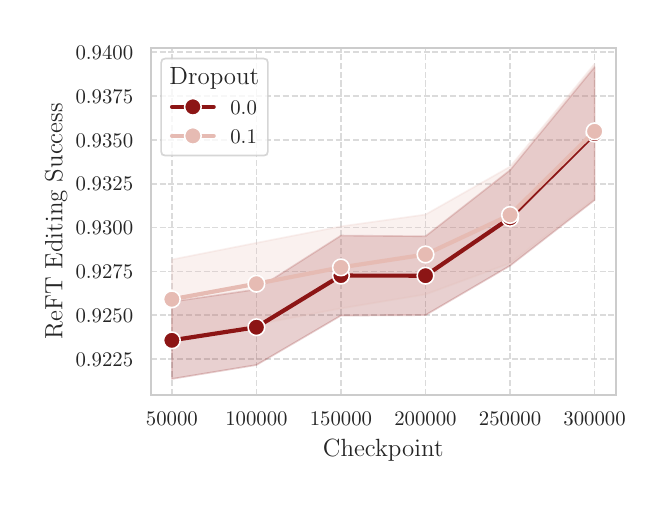}  \\
    \end{tabular}
    \caption{top: MEND edit success rate for MLM at varying levels of dropout ($(p<0.0001)$ across 5 seeds and edited concepts; margin of error are within floating point differences); bottom: ReFT edit success rate for MLM at varying levels of dropout (no significant difference across 5 seeds and edited concepts).}
    \label{fig:editing-figure}
\end{figure}

We first investigate the effects of dropout on both the pretraining objective and downstream capabilities of masked and decoder language models. To do this, we first pretrain two sets of language models with varying levels of dropout (\cref{sec:pt}) and then evaluate their capabilities through varying metrics (\cref{sec:evals}). Finally, we perform embedding and editing experiments on the pretrained BERT models to investigate the causal influence of applying dropout in pretraining (\cref{sec:Causal}). 

\subsection{Pretraining}

\label{sec:pt}

We train two sets of models: BERT-base \citep{devlin-etal-2019-bert} MLMs, and Pythia 160M and 1.4B  decoder LMs \citep{10.5555/3618408.3618510}. The BERT models are trained using the masked language modeling objective using the 10-billon-token slice of Huggingface FineWeb \citep{huggingfacefw_2024} following the optimization procedure given in \cref{sec:big-logistics}; the Pythia models are trained using The Pile Deduplicated \citep{gao2020pile} dataset (to match the original Pythia models) up to 1.5 million steps using the optimization procedure given in \cref{sec:pythia-logistics}.

We apply pretraining dropout at $p=0.0$, $p=0.1$, and $p=0.3$ on the attention and MLP blocks. We experiment with both static dropout as well as ``early dropout'' following \citet{liu2023dropout} whereby dropout is scheduled on the first $35\%$ of training before being disabled. Details on the early dropout implementation are discussed in \cref{sec:early-dropout}. 

\subsection{Capabilities Evaluations}

\label{sec:evals}
Our primary goal is to investigate how pretraining with dropout on single-epoch training affects the downstream performance of language models. To do this, we perform the following measurements:

\paragraph{(M)LM Loss} First, we measure the loss of our models on the pretraining objective on a withheld subset of the pretraining distribution. For decoder-LMs, this is simple cross-entropy loss. For MLMs, this is cross-entropy loss on the \textit{masked} training objective (i.e., the $15\%$ masked cloze task described in \cref{sec:big-logistics}). This simple measure has been widely reported as correlating with fluent natural language generation matching human-like distributions \citep{goodkind-bicknell-2018-predictive,wilcox_predictive_2020}. Hence, it serves as a simple baseline for (M)LM capabilities.

\paragraph{Morpho-syntactic Phenomena} For the decoder LMs (i.e., models that can be used in the conventional sense as a sequence ``language model''), we evaluate linguistic knowledge using the BLiMP dataset \citep{warstadt-etal-2020-blimp-benchmark}. BLiMP is an evaluation dataset of 67 different grammatical phenomena and consists of linguistic minimal pairs, exactly one of which is grammatically acceptable. We evaluate the emergence of morphology and syntax in the pretrained LMs by assessing whether the models assign a higher probability to the acceptable sequence than to the unacceptable one.

\paragraph{Fine Tuning with SQuAD} The previous two measures evaluate the pretrained LM on various tasks directly relating to the task distribution. However, LM capabilities can be unlocked through finetuning \citep{weifinetuned}. To investigate this, we fine-tune our MLMs on the SQuAD V2 dataset \citep{rajpurkar-etal-2018-know} following standard procedures described in detail in \cref{sec:squad} (in particular, \textit{with dropout enabled} tuning time regardless of whether it is on in pretraining, since we fine-tune SQuAD for more than one epoch). We report the F1 score obtained by the fine-tuned model.

\paragraph{Fine Tuning with MNLI} In addition to the SQuAD results above, we further investigate the fine-tuned capability of our MLMs on the Multi-NLI (MNLI) dataset \citep{williams-etal-2018-broad} following procedures detailed in \cref{sec:mnli}. As with SQuAD, we enable dropout in the model at a rate of 0.15 regardless of whether or not it is enabled for pretraining. We report the overall 3-class classification accuracy (``contradiction'', ``neutral'', and ``entailment'') obtained by the fine-tuned model.


\subsection{Causal and Embedding Analysis}
\label{sec:Causal}

To gain further insight into the causal influence of dropout on the language models' pretraining process, we conduct Causal embedding-level analysis on the pretrained MLMs. One lens with which to approach this investigation is to study the way LM embeddings store ``knowledge'': recent literature highlights that the storage of factual ``knowledge'' can be treated as a key-value lookup in the post-attention MLP \citep{geva-etal-2021-transformer} subject to input features; hence if dropout reduces feature co-adaptations \citep{hinton2012improving}, we hypothesize that dropout is likely to reduce the consistency with which knowledge is stored and elicited. \Cref{sec:more-motivation} provides detail on this hypothesis.

\paragraph{Formalism}
\label{sec:formal}
In this work, we define ``knowledge'' as tuples $(a,r,b)$ (``subject-relation-object''); then, if a model $M$ has ``stored'' particular knowledge, we can find some mapping $f$ that learns $f\qty(M)(a,r) = b$ but learning $f\qty(a,r) = b$ without $M$ may be difficult.

We follow the definition given in \citep{elazar-etal-2021-measuring} to impose additional constraints for human languages. We define: (1) $f$ as the cloze (mask-fill) task on the set of masked phrases (e.g. \texttt{[B] is the [R] of [A]}) which are quasi-paraphrases \citep{bhagat-hovy-2013-squibs} of each other; and (2) $M$ to be a masked-language model, in particular, a BERT.

\paragraph{Knowledge in BERT-Sized Models}
One measurable effect of consistent and localized embedding of knowledge is the increased ease of model editing. If ``knowledge'' is more localized, then one needs to edit a smaller area of the MLP to corrupt or change it.

Therefore, we measure the effects that pretraining with and without dropout has on the editability of downstream models. We define ``editing'' here as changing a pattern (i.e. $(a,r,b)$ to $(a,r,b')$ with $b \neq b'$) under all choices $H$. We obtain these patterns from the Pararel \citep{elazar-etal-2021-measuring} dataset, which consists of varying syntactic frames for expressing the same relational concept. We randomly rearrange objects within each relation category and edit the model to generate the newly paired object for each relation. We then evaluate the model's prediction test accuracy on masked tokens with a balanced mix of permuted and unrelated relations. 

We train two model editing techniques: Representation Fine Tuning (ReFT) \citep{wu2024reft} and MEND \citep{mitchellfast}. Further details of these approaches are given in \cref{sec:reft} and \cref{sec:mend} respectively.


\section{Results}

\paragraph{Removing dropout makes more capable models} Across all evaluations of capabilities described in \cref{sec:evals}, models pretrained with dropout, \textit{including early dropout}, performed worse than those pretrained without dropout. As \cref{fig:large-figure} shows, the mean LM loss of the models trained without dropout is lower than those trained with any dropout. The models trained without dropout also scored higher on both BLiMP, and SQuAD F1 while performing marginally better on MNLI accuracy. Furthermore, we discover that dropout rate correlates with its effect size on LM performance; we discuss this effect in \cref{sec:sweep}.

\paragraph{...that are slightly more editable}
As shown in \cref{fig:editing-figure}, given sufficient training steps, the model trained without dropout is statistically more successful in editing using MEND and did not have notable editing differences by ReFT. \Cref{sec:reft-v-mend} discusses this result in detail: we believe applying dropout yielded little difference in ReFT performance because ReFT edits orthogonal subspaces, meaning the \textit{distance} between stored knowledge disturbed by dropout is less important.



\section{Conclusion and Discussion}
Although it has been standard practice to apply dropout to regularize neural network training, we find---consistent with the recent trend of removing dropout in massive LM pretraining---that dropout in a single-epoch pretraining regime is not necessary and hurts performance. In particular, we find that the models pretrained \textit{without} dropout score consistently more favorably in capability measures of LM loss, morpho-syntax generalization in BLiMP, and fine-tuned SQuAD. These results hold even with the early-dropout technique \citep{liu2023dropout}. Furthermore, we note that the models trained without dropout improve in editability after training with sufficient scale with MEND\@. We hypothesize that this is because dropout non-discriminately limits co-adaptation in input features \citep{hinton2012improving}, leading to less localized representations of ``knowledge'', resulting in multiple independent copies of facts stored. This ultimately makes models less amenable to editing.

Taken together, our conclusions indicate that, with single epoch LM pretraining, one should \textbf{drop dropout}.


\section{Limitations}

\paragraph{Statistical-Theoretical Grounding} Though this work provides a strong empirical framework and evidence for removing dropout, it attempts to make no theoretical claims about the expected behavior of dropout. Such an evaluation is difficult naively because the expected value of dropout converges to the identity \citep{baldi2013understanding}. However, if taking a lens of dropout as a regularizer in the first order \citep{wager2013dropout}, this would be a fruitful avenue for future work which is made possible to validate by the empirical evidence here.

\paragraph{Scaling Laws} The emergence of overfitting scales with the parameter count of the model. Due to resource limitations (i.e. the fact that this experimental setup requires pretraining multiple seeds of an LM), it is difficult to directly measure how the result scales. However, given the replication from MLMs and decoder LMs ranging from 160M to 1.4B parameters, and the previously reported effects of dropout \textit{strengthen} as models scale \citep{liu2023dropout}, we believe these results are applicable for pretraining in similar settings.

\paragraph{Activation and ROME-style Probes} We elected to use neither integrated-gradient style activation probes nor causal corruption probes in this work due to recent evidence that those probes are fairly input-dependent, not easily localized, and not able to cleanly probe stored ``knowledge'' in a network \citep{niu2024does,hase2024does}. 










\section*{Acknowledgements}
We would like to thank our colleagues across Stanford NLP for their insights. In particular, we would like to acknowledge Róbert Csordás, Shikhar Murty, Amelia F. Hardy, Julie Kallini, Liam Kruse, and Ethan Hsu for their valuable time and ideas. We would finally like to thank the anonymous reviewers and editors for their gracious feedback. Houjun Liu was supported in part by the Stanford CURIS program during this work. Christopher D.~Manning is a CIFAR fellow. 

\bibliography{anthology,custom}
\appendix

\section{Motivating the Studying of Dropout Through Knowledge Storage}
\label{sec:more-motivation}
The central claim of \citet{hinton2012improving} is that dropout reduces feature co-adaptations. Framing dropout under this lens implies that dropout would modulate the way knowledge is elicited under permutation of the selection of $h$. The application of dropout would lead the model to learn multiple independent pathways to represent $a,r$ instead of relying on co-adapted features across all existing $h(a,r)$. While this implies our model may be more robust to previously unseen $h'$ that do not share the co-adapted features of the training set, it could also then result in multiple, independent representations of $a,r$ being built under the choice of $h_j$---leading to possible inconsistency. 

Specific to human language modeling, the former lens (dropout leads to more robust models under input sequence permutation) would imply that a model that uses dropout would be more consistent and less likely to hallucinate---a result that has been discussed in literature \citep{agarwal2018hallucinations}; the latter lens would imply that a model which uses dropout would build unrelated representations of knowledge that may not be consistent---a result that also has been discussed \citep{elazar-etal-2021-measuring}. This work seeks to resolve this tension between two lenses.

\section{BERT Pretraining Details}
\label{sec:big-logistics}
We used the Huggingface \citep{wolf-etal-2020-transformers} implementation of BERT-Large \citep{devlin-etal-2019-bert}, and only modified it to disable attention and MLP dropout globally. When MLP dropout is used, $p=0.1$. Optimization was done \textit{with regularization} using AdamW \citep{loshchilov2017decoupled} following published parameters:

\begin{table}[ht]
    \centering
    \small
    \begin{tabular}{@{}rl@{}} \toprule
        Parameter & Value  \\\midrule
        LR & linear warmup 10k, linear decay, peak $1\times 10^{-4}$ \\
        Adam $\beta$ & $$(0.9, 0.999)$$ \\
        Adam $\epsilon$& $1 \times 10^{-6}$\\\bottomrule
    \end{tabular}
    \caption{Details of the Small Scale Model}
    \label{tab:small_structure_params}
\end{table}

The model was trained on the officially sampled 10BT slice of Huggingface FineWeb \citep{huggingfacefw_2024}, running with fully-sharded data-parallel over 4 GPUs for a joint batch size of $4 \times 96 = 384$. Batching was done sequentially with the Pytorch Data Loader, sequence lengths are capped at $512$ tokens. The pretrained tokenizer for BERT-large available on Huggingface was used instead of training a tokenizer from scratch; sequences are wrapped with usual start and end sequence tokens. The model was optimized over 8 days on Nvidia A100 GPUs.

The modeling objective was a span-corruption loss, which uses official sampling rates for incidences of corruption and cloze. Tokens are selected for corruption with $15\%$ chance, and of which $10\%$ are shuffled, $80\%$ are masked, and the rest are returned normally. Sampling is done dynamically with a fixed seed for each run. 

\section{Pythia Pretraining Details}
\label{sec:pythia-logistics}

As with \cref{sec:big-logistics}, we used the Huggingface \citep{wolf-etal-2020-transformers} implementation of the Pythia suite of models \citep{10.5555/3618408.3618510}; the only modifications we performed involves modulating attention and MLP dropout globally as needed for each experiment. Optimization was done \textit{with regularization} using AdamW \citep{loshchilov2017decoupled} following published parameters:

\begin{table}[ht]
    \centering
    \small
    \begin{tabular}{@{}rl@{}} \toprule
        Parameter & Value  \\\midrule
        LR & linear warmup 10\%, linear decay, peak $6\times 10^{-4}$ \\
        Adam $\beta$ & $$(0.9, 0.999)$$ \\
        Adam $\epsilon$& $1 \times 10^{-6}$\\\bottomrule
    \end{tabular}
    \caption{Details of the Small Scale Model}
    \label{tab:pythia_small_structure_params}
\end{table}

We trained both the 160M and 1.4B variants of the Pythia models on the first 1.5 million steps (around 1.1BT) of The Pile Deduplicated \citep{gao2020pile} dataset, running with fully-sharded data-parallel over 2 GPUs for a joint batch size of $2\times 16 = 32$ for the smaller variant and $2 \times 2 = 4$ for the larger variant. Batching was done in local-shuffle random order with a buffer size of $1,000$, and sequence lengths are truncated to the models' maximum length, being $2048$. The pretrained tokenizer for each of the models were used instead of training a tokenizer on scratch. The model was optimized over 8 days on Nvidia A100 GPUs.

The model objective used was standard cross-entropy language modeling loss.

\section{Early Dropout Implementation}  
\label{sec:early-dropout}
We used a simple binary early dropout schedule described in Appendix D of \citep{liu2023dropout} in addition to the training parameters of the Pythia suite of models described in \cref{sec:pythia-logistics}. In particular, we disabled dropout at $35\%$ of training as a hard cutoff, having trained the model for 77,821 of roughly 220,000 steps of the training corpus. No other training parameters (including optimizer states) are reset after disabling dropout, and training continues as usual.

\section{MEND Implementation}
\label{sec:mend}
MEND \citep{mitchellfast} is a model editing technique that projects the fine-tuning gradient of a model to weight updates that result in well-localized edits (i.e. edits which do not affect non-edited facts). For target knowledge to store $(a,r,b)$ and unrelated knowledge which we don't want changed $(a', r', b')$, it does this by learning a function $f: U(a,r) \times \Delta(a,r,b) \to \nabla W$ that takes the pre-MLP activations $u(a,r)$ and post-MLP fine-tuning gradient for the layer $\delta(a,r,b)$ and produces the appropriate MLP weight difference $\nabla W$. To learn $f$a, we optimize for an objective which minimizes a joint loss:

\begin{equation}
    L = c_{edit} L_{edit} + L_{loc}
\end{equation}

whereby $L_{edit}$ is the negative log-likelihood of the desired post-edit token $-\log p_{post\ edit}(b|a,r)$ and $L_{loc}$ is the KL-divergence of the posterior distribution of the model against reference $KL(p_{ref}(b'|a', r')||p_{post\ edit}(b'|a', r'))$.

$f$ is an identity-like projection that takes a vector in $V = U(a,r) \times \Delta(a,r,b)$ (the pre-layer activation concatenated with the post-layer fine-tuning gradient) and maps it in the following manner:

\begin{align}
    h(v) &= ReLU(U_1V_1\  norm(v)) + norm(v)\\
    f(v) &= ReLU(U_2V_2\ h(v)) + h(v)
\end{align}

notably, $V_j$ are Xavier initialied while $U_j$ is zero-initialised, making this function $f$ the identity prior to tuning. $U_j, V_j$ is learned.

To learn the edit, we shuffle the targets for each ``knowledge'' given in the Pararel patterns \citep{elazar-etal-2021-measuring} (i.e. for tuple \texttt{(a,r,b)}, we switch \texttt{b} to something else sampled in the dataset). We then split the patterns (i.e. $h$ in the formalism given in \cref{sec:formal}, the quasi-rephrasing) into a $95\%-5\%$ train-test split for each knowledge. Edits are learned from all train splits at once, and are tested on all test splits at once to report edit success. Only the accuracy on the target token is reported.

We learn the edits using a learning rate of $5\times10^{-5}$, with a batch size of $12$ and an Adam optimizer. Optimization was done over 3 hours on Nvidia RTX a6000 Ada Generation GPUs. The pretrained models on which the edits were performed did not have dropout on, regardless of whether they are pretrained with dropout.

\section{ReFT Implementation}
\label{sec:reft}

ReFT is a model editing technique that intervenes on a certain number of prefix and suffix tokens (i.e. the first and last \texttt{n} tokens') embeddings of a model by perturbing their embeddings in an orthogonal subspace to help localize the edit (i.e. ensuring that there is nothing else that is influenced by the edit). In particular, it learns linear projections weight $W$, and bias $b$ which is applied to the post-attention hidden projection of the editing layer following:

\begin{equation}
    E(h) = h + R^\top \qty(Wh + b - Rh)
\end{equation}

whereby $R$ is a matrix with enforced orthonormal rows. 

We learn a rank-4 edit on layers 4, 6, 8, and 11, intervening on one prefix and suffix tokens out of each sequence. We train the intervention one concept at a time and evaluate on $10\%$ of the held-out patterns.

We train the procedure on one concept at a time: we want to isolate the parameters necessary to only perform the cloze task correctly for that concept; note that this includes normal span corruption on non-concept tokens such as stop words, so properties of general language modeling is not lost. 

Edit success is measured by mask token match overall concepts. Optimization was done for one epoch with a learning rate of $5\times10^{-4}$ with the Adam optimizer, with batch sizes of $10$ patterns at a time, which took roughly 12 minutes to train for each concept on Nvidia RTX a6000 Ada Generation GPUs. Dropout was disabled during edit and evaluation regardless of whether pretraining used dropout.

\section{SQuADV2 Fine Tuning}
\label{sec:squad}

We trained on the train slice of SQuAD v2 \citep{rajpurkar-etal-2018-know} for 2 epochs using the Adam optimizer, at a learning rate of $1 \times 10^{-5}$ with a batch size of $12$. Dropout rate was set to $0.1$ during training and evaluation regardless of whether pretraining used dropout as is reported in \citep{devlin-etal-2019-bert}. An adapted version of the official evaluation script was used to obtain the dev-slice results reported in this work. Questions and answers are separated by the \texttt{[MASK]} token, which was previously used for MLM in the pretraining. A randomly initialized sequence prediction head is added on top of the pretrained network. Non-answers are represented by the model predicting the null span (i.e. starting at \texttt{[CLS]} and ending at \texttt{[CLS]}---the starts of sequences).

\section{MNLI Fine Tuning}
\label{sec:mnli}

We fine-tune our trained BERT models on the train slice of the original MNLI \citep{williams-etal-2018-broad} dataset available \footnote{https://huggingface.co/datasets/nyu-mll/multi\_nli} on Huggingface for 3 epochs using the adam optimizer, at a fixed learning rate of $2 \times 10^{-5}$. Batch size was set to $128$, and dropout rate was set to $0.15$ regardless of whether pretraining the BERT model used dropout consistent with previous approaches. Premise and hypotheses were separated by the \texttt{[SEP]} token, and the final token residual was decoded using a single MLP to form a three-class classification model, which is initially randomly initialized.

\section{Performance Differences in ReFT vs. MEND}
\label{sec:reft-v-mend}
In this work, we found that while MEND had significantly higher editing success in the no dropout case after sufficient training, ReFT's edit success simply converged. We believe the relative higher degree of success in the no-dropout case for MEND is expected: ReFT is formulated to only edit on orthogonal subspaces \citep{wu2024reft}, meaning the Euclidean distance between embedding clusters is less important; however, the convergence indicates that the no-dropout model was indeed still able to build equivalently useful embeddings for edits like ReFT. 

\section{Sweeping Dropout Rates}
\label{sec:sweep}

\begin{figure}[ht]
    \centering
    \hspace{-0.01pt}
    \begin{tabular}{c}
      \includegraphics[width=0.45\textwidth,trim={0.7cm 0.7cm 0.7cm 0.7cm},clip]{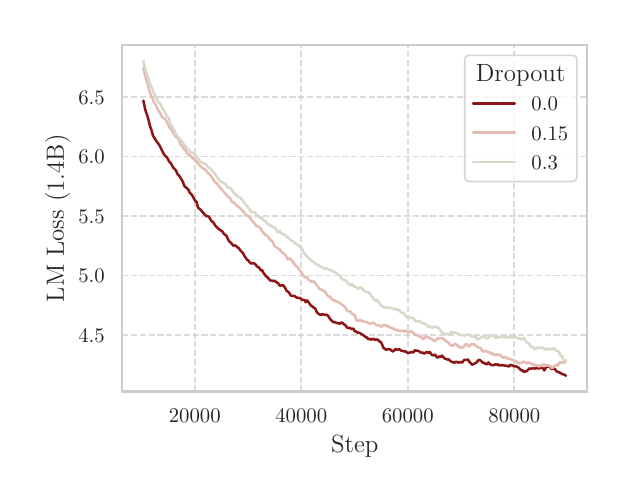}  \\
    \end{tabular}
    \caption{Sweeping dropout rate on the 1.4B parameter Pythia model, across $p=0.0$, $p=0.15$, $p=0.3$. Loss averaged across 3 seeds.}
    \label{fig:sweep}
\end{figure}

As seen in \cref{fig:sweep}, the effects of dropout presents as a function of the rate of dropout. In particular, while applying lower amounts of dropout (such as $p=0.15$) results in performance gap less dramatic than applying the originally proposed $p=0.3$, the model nevertheless learns slower and converges less stably. The decreased rate of converge is roughly related to the amount of dropout applied, which is a finding consistent with the investigations of late dropout in \citet{liu2023dropout}.

\end{document}